\documentclass[sigconf,nonacm]{acmart}

\AtBeginDocument{%
  \providecommand\BibTeX{{%
    \normalfont B\kern-0.5em{\scshape i\kern-0.25em b}\kern-0.8em\TeX}}}

\begin{document}

\title{Current and Future Challenges in Humanoid Robotics - An Empirical Investigation}

\author{Maike Paetzel-Prüsmann}
\email{me@maike-paetzel.de}
\affiliation{%
  \institution{Independent Researcher, Switzerland}
  \country{}
}

\author{Alessandra Rossi}
\authornote{This work has been supported by the Italian PON R\&I 2014-2020 - REACT-EU (CUP E65F21002920003)}
\email{alessandra.rossi@ieee.org}
\affiliation{%
  \institution{University of Naples Federico II, Italy}
  \country{}
}

\author{Merel Keijsers}
\email{mkeijsers@johncabot.edu}
\affiliation{%
  \institution{John Cabot University, Italy}
  \country{}
}

\settopmatter{printacmref=false, printccs=true, printfolios=true}

\begin{abstract}
  The goal of RoboCup is to make research in the area of robotics measurable over time, and grow a community that works together to solve increasingly difficult challenges over the years. The most ambitious of these challenges it to be able to play against the human world champions in soccer in 2050. To better understand what members of the RoboCup community believes to be the state of the art and the main challenges in the next decade and towards the 2050 game, we developed a survey and distributed it to members of different experience level and background within the community. We present data from 39 responses. Results highlighted that locomotion, awareness and decision-making, and  robustness of robots are among those considered of high importance for the community, while human-robot interaction and natural language processing and generation are rated of low in importance and difficulty.
\end{abstract}

\keywords{Robotics, Humanoid Robots, Empirical Study, RoboCup, Robot Soccer}

\maketitle

\section{Introduction}
Over the last decades, robots have reached substantial improvements in their robustness, precision, actuation speed, and number of tasks they are reliably able to perform. In almost all industrial production lines, robots have become an integral part of the process \cite{SINGH20221779}. When it comes to our daily lives, however, robots are still sparse. In home environments, only vacuum cleaning and lawn mowing robots have been widely adopted. In the service industry robots are still absent apart from a few notable exceptions, such as robot hotels or restaurants \cite{10.1145/3434074.3447206,foodbot1}. Unlike in industrial settings, these placements usually happen under close monitoring by human staff members. 

Measuring and quantifying the research achievements in robotics is challenging. The RoboCup Federation\footnote{The RoboCup \url{https://www.robocup.org/}} has established one approach, which is a series of well-defined tasks to be performed by robots, which increase in difficulty every year. These tasks are distributed into different leagues, ranging from logistics applications to domestic environments and rescue scenarios. One area of particular interest is robot soccer, which requires a multitude of integrated skills to be performed in real time and with high precision. The goal of the RoboCup soccer leagues is to achieve a level of skill for their robots to be able to challenge the human world soccer champions in 2050 and win the game. For the match to be fair, many consider requirements for robots to have humanlike locomotion, sensing and communication to be integral. These constraints are particularly observed by the RoboCup Humanoid League, which is largely seen as the one to work most diligently towards the 2050 goal. The league has developed a roadmap to increase challenges for their robots with the 2050 goal in mind. For example, the field size is still small and needs to substantially grow in order to reach the field size of human soccer matches. 

In this paper, we examine what the people involved in RoboCup competitions believes the major challenges of humanoid robotics to be, both for the next decade and with the 2050 goal in mind. We also want to understand whether roboticists themselves believe the 2050 goal to be achievable, and if so, what the most important tasks will be. To this extent, we developed a survey, and distributed it among teams participating in RoboCup. In the following, we describe the questionnaire that we designed in details, and we summarize the results from the 39 people who participated in our survey.

\section{Methodology}

An online survey was designed using LimeSurvey and the link was sent out to the general RoboCup mailinglist and all individual RoboCup league mailing lists. 
The questionnaire consisted of three main parts. First, participants were asked to provide some demographic details, such as for how many years they participated in RoboCup, which leagues they have experience in and what their current study level or job title is. We also asked them to describe their own background and research focus in a free-text field. In the second part, we wanted to understand what they believe to be challenges and important focuses for RoboCup research on the road to 2050. We first asked them in a free-text field to describe what they believe to be the biggest challenges for research in the area of humanoid robotics in the next decade, and to achieve the goal of 2050. Then, we provided them with 12 research areas and asked them to rank them according to how important they believe they will be to reach the 2050 goal, and in a second ranking how difficult they believe achieving a level suitable for the 2050 game will be (see Table \ref{tab:ranking}). In the last set of questions, we asked participants when they believe the first full game between humanoid robots and humans will take place, and what they believe to be the final score in 2050.

In total, 39 people participated in our survey, and they stated to be participating in or being associate with RoboCup. We were able to recruit a wide variety of experiences within RoboCup, with an average of 9.51 years of involvement in RoboCup ($SD = 6.74$). 17 participants stated to be students (6 Bachelor, 6 Master, 5 PhD level), 2 participants are Postdoctoral researchers, 1 is a researcher from outside Academia, 2 participants are academic researchers, and 12 work as professors (3 Assistant, 2 Associate, 7 Full Professor).  5 participants indicated "other" as their job title, most of them specified to be industry professionals.  Our participants have experience in all the RoboCup leagues,  with 16 of them coming from the Humanoid league (HL),  12 participants with experience in 2D or 3D simulation leagues,  10 participants from the Standard Platform League (SPL) and Small Size League,  7 participants from RoboCup Rescue,  5 participants from any of the @home leagues,  4 participants in both Middle Size and RoboCup Junior leagues, 2 participants in Logistics and 1 in @Work.

\section{Results}

\begin{table}
    \centering
    \caption{Research topics ranked by participants for their importance and difficulty to achieve a suitable level for the 2050 games.}
    \begin{tabular}{l | c | c}
        Research Topic & Rank Importance  & Rank Difficulty \\
        Locomotion & 1  & 1 \\
        Awareness & 2  & 4 \\
        Robustness & 3  & 2 \\
        Decision-making & 4 & 3 \\
        Safety & 5 & 5 \\
        Actuators & 6 & 8 \\ 
        Robot-Robot Collab. & 7 & 9 \\
        Computer Vision & 8 & 11 \\
        Human-Robot Interaction & 9 & 6 \\
        Energy  & 10 & 7 \\
        Surface Material & 11 & 12 \\
        Natural Language & 12 & 10
    \end{tabular}
    \label{tab:ranking}
\end{table}
We first coded the free-text responses to the question of what participants believe to be the biggest challenges in robotics research in the next decade using the same 12 research areas that were provided to them in the next step. At this first question which was entirely unbiased by the research areas provided by us, locomotion ($n=11$), robustness ($n=7$) and computer vision ($n=5$) were the topics mentioned most often by participants. Safety, surface material and natural language processing and generation were not brought up at all. The remaining comments that could not be subsumed under one of our codes were concerned with integration, operating in different environmental conditions, artificial intelligence and the costs involved with building one or multiple robots. Then, we coded the free-text responses to the question about the biggest challenges to achieve the 2050 goal. We observed again that locomotion appeared most often in participants responses ($n=9$), while robustness ($n=3$) and computer vision ($n=4$) were the less frequently brought up. Energy resources and consumption ($n=7$) accompany by safety and decision-making ($n=5$) were mentioned more often than in the previous question. The surface material was, again, not directly brought up at all. Other responses that did not fit into one of our pre-determined categories could mainly be summarized as concerns about costs, and general Artificial Intelligence (AI) capabilities that are required.

These results are in line with the ranking that participants provided both in the importance of research topics for reaching the 2050 goal and in the difficulty of achieving a suitable level for the 2050 games (see Table \ref{tab:ranking}). Locomotion was ranked as both the most important and most difficult research topic. It is followed by awareness of the environment and situation (2 in importance, 4 in difficulty), robustness (3 in importance, 2 in difficulty) and decision-making (4 in importance, 3 in difficulty). Almost all topics have a similar rank in importance and difficulty. The ones that differ most are computer vision (8 in importance, 11 in difficulty), Human-Robot Interaction (9 in importance, 6 in difficulty) and energy resources and consumption (10 in importance, 7 in difficulty). 

When asked in which year the first full game between humans and robots will take place, the responses ranged between 2025 and 2100, with a median of 2040. One participant wrote 9999, which we interpret as they do not believe this game to happen at all. Two participants noted dates in the past. If we have a game between humans and robots in 2050, 27 participants believe this game will be won by the human players. In case the humans win, the average estimated goal difference between the teams is 12.89 with a median of 5.  Eight participants believe the robots will win, in which case they would win with an average goal difference of 3.5, median of 2.5. The remaining four participants believe the game will end in a draw. Only two participants believe the humans will not be able to score a single goal, while 14 do not believe the robots will be able to score at all in the 2050 match.

\section{Discussion \& Conclusion}
In this paper, we analyzed results from a survey distributed to different leagues within the RoboCup community to capture the impression of researchers in different areas of robotics about the biggest challenges in robotics research in the next decade, and with the 2050 goal of RoboCup in mind. The areas rated as most important and challenging are all areas that are of active research interest at the moment. On the contrary, topics with less prominent research focus to date, like human-robot interaction and natural language processing and generation, are rated as low in importance and difficulty. This potentially shows that researchers believe other problems to be more imminent and pre-conditions that needs to be at a much more robust level before moving to these research areas. It does, however, also show that little attention is given these areas, which may still turn out to be of importance for the 2050 games to be a success. Interestingly, the sample of researchers who responded to our survey does largely not believe the robots to be able to win the match in the year 2050. However, they predict the first full game between humans and robots to take place in 2040, which is only two decades after this survey was conducted and one decade before the goal of 2050.

\bibliographystyle{ACM-Reference-Format}
\bibliography{biblio}

\end{document}